\documentclass{ecai}
\usepackage{times}
\usepackage{graphicx}
\usepackage{latexsym}

\usepackage{times}
\usepackage{float}
\usepackage{helvet}
\usepackage{courier}
\usepackage{blindtext, graphicx}
\usepackage[utf8]{inputenc}
\usepackage{listings}
\usepackage{graphicx}
\usepackage{subcaption}
\usepackage{amsthm}
\usepackage{hyperref}
\usepackage{enumitem}
\usepackage{algorithm, algpseudocode}
\usepackage{caption}
\usepackage{amsmath}
\usepackage{lipsum}
\usepackage{todonotes}

\theoremstyle{plain}
\theoremstyle{definition}
\newtheorem{definition}{Definition}

\newcommand\pred[1]{\texttt{\footnotesize{#1}}}

\usepackage{listings}
\begin{document}

\title{Landmark-Based Plan Recognition\footnote{This document is a full paper of a work published (as short paper) in the 22nd European Conference on Artificial Intelligence (ECAI), 2016.}}

\author{Ramon Fraga Pereira$^2$ \and Felipe Meneguzzi \institute{Pontifical Catholic University of Rio Grande do Sul (PUCRS), Brazil. Contact: ramon.pereira@acad.pucrs.br and felipe.meneguzzi@pucrs.br} }

\maketitle
\bibliographystyle{ecai}


\begin{abstract}
Recognition of goals and plans using incomplete evidence from action execution can be done efficiently by using planning techniques. 
In many applications it is important to recognize goals and plans not only accurately, but also quickly.
In this paper, we develop a heuristic approach for recognizing plans based on planning techniques that rely on ordering constraints to filter candidate goals from observations.
These ordering constraints are called landmarks in the planning literature, which are facts or actions that cannot be avoided to achieve a goal.
We show the applicability of planning landmarks in two settings: first, we use it directly to develop a heuristic-based plan recognition approach; second, we refine an existing planning-based plan recognition approach by pre-filtering its candidate goals. 
Our empirical evaluation shows that our approach is not only substantially more accurate than the state-of-the-art in all available datasets, it is also an order of magnitude faster.
\end{abstract}

\section{Introduction}

As more computer systems require reasoning about what agents (both human and artificial) other than themselves are doing, the ability to accurately and efficiently recognize goals and plans from agent behavior becomes increasingly important. 
Plan recognition is the task of recognizing goals and plans based on often incomplete observations that include actions executed by agents and properties of agent behavior in an environment~\cite{ActivityIntentPlanRecogition_Book2014}. 
Accurate plan recognition is important to monitor and anticipate agent behavior, such as in crime detection and prevention, monitoring activities, and elderly-care. 
Most plan recognition approaches~\cite{Geib_ProbabilisticPlanRecognition_MOO2005,AvrahamiZilberbrandK_IJCAI2005} employ plan libraries (\textit{i.e}, a library with all plans for achieving a set of goals) to represent agent behavior, resulting in approaches to recognize plans that are analogous to language parsing. 
Recent work~\cite{RamirezG_IJCAI2009,RamirezG_AAAI2010,PattisonGoalRecognition_2010,NASA_GoalRecognition_IJCAI2015} use planning domain definitions (domain theories) to represent potential agent behavior, bringing plan recognition closer to planning algorithms. 
These approaches allow techniques used in planning algorithms to be employed for recognizing goals and plans.

In this paper, we develop a plan recognition approach that relies on planning landmarks~\cite{Porteous_Landmarks_2002,Hoffmann2004_OrderedLandmarks} to filter candidate goals and plans from the observations. 
Landmarks are properties (or actions) that every plan must satisfy (or execute) at some point in every plan execution to achieve a goal. Whereas in planning algorithms these landmarks are used to focus search, in our approach, they allow plan recognition algorithms to rule out candidate goals whose landmarks are missing from observations. 
Thus, based on planning landmarks, we develop an algorithm to filter candidate goals by estimating how many landmarks required by every goal in the set of candidate goals have been achieved within the observed actions.  
Since computing a subset of landmarks for a set of goals can be done very quickly, our approach can provide substantial runtime gains. 
In this way, we use this filtering algorithm in two settings. 
First, we build a landmark-based plan recognition heuristic that analyzes the amount of achieved landmarks to estimate the percentage of completion of each filtered candidate goal. 
Second, we show that the filter we develop can also be applied to other planning-based plan recognition approaches, such as the approach from Ram{\'{\i}}rez and Geffner~\cite{RamirezG_IJCAI2009}.

We evaluate empirically our plan recognition approach against the current state-of-the-art~\cite{RamirezG_IJCAI2009} by using openly available datasets for plan recognition developed by 
Ram{\'{\i}}rez and Geffner in~\cite{RamirezG_IJCAI2009,RamirezG_AAAI2010}, and which have been used to evaluate recent approaches to plan recognition~\cite{NASA_GoalRecognition_IJCAI2015}.
This dataset provides several domains and problems in which it is not trivial to recognize the intended goal from a set of candidate goals and observations.
Using this dataset, we show that our approach has at least three advantages over existing approaches. 
First, by relaxing the filter using a small threshold our landmark-based plan recognition approach is more accurate than the current state-of-the-art~\cite{RamirezG_IJCAI2009}. 
Second, our approach also provides substantially faster recognition time on its own and when used to improve existing plan recognition approaches. 
Finally, we show that our filtering algorithm provides substantial improvements in recognition time when used to improve existing plan recognition approaches.

This paper is organized as follows. 
Section~\ref{section:Background} provides background on planning and plan recognition. 
In Section~\ref{section:InferringStructures}, we describe how we extract useful information from planning domain definition. Sections~\ref{section:FilteringGoals},~\ref{section:HeuristicPlanRecognition}, and~\ref{section:landmarkPlanRecognition} develop the key parts of our approach for plan recognition. 
We empirically evaluate our approach in Section~\ref{section:ExperimentsAndEvaluation}, which shows the results of the experiments. 
In Section~\ref{section:RelatedWork}, we survey related work and compare the state of the art with our approach. 
Finally, in Section~\ref{section:Conclusion}, we conclude this paper by discussing limitations, advantages and future directions of our approach. 

\section{Background}\label{section:Background}

In this section, we provide essential background on planning terminology, and how we define plan recognition problems over planning domain definitions.

\subsection{Planning}

Planning is the problem of finding a sequence of actions (\textit{i.e}, plan) that achieves a particular goal from an initial state. 
In this work, we adopt the terminology from Ghallab~\emph{et al.}~\cite{AutomatedPlanning_Book2011} to represent planning domains and problems. First, we define a state in the environment by the following Definition~\ref{def:state}. 

\begin{definition} [\textbf{Predicates and State}]\label{def:state}
\textit{A predicate is denoted by an n-ary predicate symbol $p$ applied to a sequence of zero or more terms ($\tau_1$, $\tau_2$, ..., $\tau_n$) -- terms are either constants or variables.
A state is a finite set of grounded predicates (facts) that represent logical values according to some interpretation. 
Facts are divided into two types: positive and negated facts, as well as constants for truth ($\top$) and falsehood ($\bot$).}
\end{definition}

\begin{definition} [\textbf{Operator}]\label{def:operator}
\textit{An operator $a$ is represented by a triple $\langle$\textit{name}($a$), \textit{pre}($a$), \textit{eff}($a$)$\rangle$: \textit{name}($a$) represents the description or signature of $a$; \textit{pre}($a$) describes the preconditions of $a$, a set of predicates that must exist in the current state for $a$ to be executed; \textit{eff}($a$) represents the effects of $a$. 
These effects are divided into \textit{eff}($a$)$^+$ (\textit{i.e}, an add-list of positive predicates) and \textit{eff}($a$)$^-$ (\textit{i.e}, a delete-list of negated predicates).}
\end{definition}

A plain domain contains operator definitions, which represents the environment dynamics that guide an agent's search for plans to achieve its goals. 
Operator definitions are used in the construction of a planning domain, which represents the environment dynamics that guide an agent's search for plans to achieve its goals.
An agent can modify the current state by executing actions according to Definition~\ref{def:action}.

\begin{definition} [\textbf{Action}]\label{def:action}
\textit{An action is a ground operator instantiated over its free variables. 
Thus, if all operator free variables are substituted by objects when instantiating an operator, we have an action.}
\end{definition}

\begin{definition}[\textbf{Planning Domain}]\label{def:planningDomain}
\textit{A planning domain definition $\Xi$ is represented by a pair $\langle \Sigma, \mathcal{A} \rangle$, which specifies the knowledge of the domain, and consists of a finite set of facts $\Sigma$ and a finite set of actions $\mathcal{A}$.}
\end{definition}

A planning instance, comprises both a planning domain and the elements of a planning problem, describing the initial state of the environment and the goal which an agent wishes to achieve as formalized in Definition~\ref{def:planningInstance}.

\begin{definition} [\textbf{Planning Instance}]\label{def:planningInstance}
\textit{A planning instance $\Pi$ is represented by a triple $\langle \Xi, \mathcal{I}, G\rangle$, in which $\Xi =  \langle \Sigma, \mathcal{A}\rangle$ is the domain definition; $\mathcal{I} \subseteq \Sigma$ is the initial state specification, which is defined by specifying the value for all facts in the initial state; and $G \subseteq \Sigma$ is the goal state specification, which represents a desired state to be reached.}
\end{definition}

Classical planning representations often separate the definition of $\mathcal{I}$ and $G$ as part of a planning problem (to be used together with a domain $\Xi$). Finally, a plan is the solution to a planning problem, as formalized in Definition~\ref{def:plan}.

\begin{definition} [\textbf{Plan}]\label{def:plan}
\textit{A plan $\pi$ for a plan instance $\Pi = \langle \Xi, \mathcal{I}, G\rangle$ is a sequence of actions $\langle$$a_1$, $a_2$, ..., $a_n$$\rangle$ that modifies the initial state $\mathcal{I}$ into one in which the goal state $G$ holds by the successive execution of actions in a plan $\pi$. 
A plan $\pi^{*}$ with length $|\pi^{*}|$ is optimal if there exists no other plan $\pi'$ for $\Pi$ such that $\pi' < \pi^{*}$.}
\end{definition}

\subsection{Plan Recognition}

Plan recognition is the task of recognizing how agents achieve their goals by observing their interactions in an environment~\cite{ActivityIntentPlanRecogition_Book2014}. 
In plan recognition, such observed interactions are defined as available evidence that can be used to recognize plans. 
Most plan recognition approaches require knowledge of an agent's possible plans for representing its typical behavior, in other words, this knowledge provides the recipes (\textit{i.e}, know-how) for achieving goals. 
These recipes are often called plan libraries and are used as input for many plan recognition approaches~\cite{Geib_ProbabilisticPlanRecognition_MOO2005,AvrahamiZilberbrandK_IJCAI2005}. 
However, in this work we use as input a planning domain definition, more specifically, we use the STRIPS~\cite{STRIPSFikes1971} fragment of PDDL~\cite{PDDLMcdermott1998}. 
We follow Ramírez and Geffner~\cite{RamirezG_IJCAI2009,RamirezG_AAAI2010} to formally define a plan recognition problem over a planning domain definition as follows.

\begin{definition}[\textbf{Plan Recognition Problem}]\label{def:planRecognition}
\textit{A plan recognition problem is a quadruple $T_{PR}$ $=$ $\langle\Xi,\mathcal{I} ,\mathcal{G}, O\rangle$, in which $\Xi$ $=$ $\langle$$\Sigma$, $\mathcal{A}$$\rangle$ is the domain definition, and consists of a finite set of facts $\Sigma$ and a finite set of actions $\mathcal{A}$; $\mathcal{I}$ represents the initial state; $\mathcal{G}$ is the set of possible goals of a hidden goal $G$, such that $G$ $\in$ $\mathcal{G}$; and 
$O$ $=$ $\langle$$o_1$, $o_2$, ..., $o_n$$\rangle$ is an observation sequence of a plan execution with each observation $o_i \in O$ being an action in the finite set of actions $\mathcal{A}$ from the domain definition $\Xi$.  This observation sequence can be full or partial, which means that for a full observation sequence we observe all actions during the execution of an agent plan, and for a partial observation sequence, only a sub-sequence of actions of the execution of an agent plan is observed. The solution for this problem is to find a hidden goal $G$ in the set of possible goals $\mathcal{G}$ that the observation sequence of a plan execution achieves.}
\end{definition}


\section{Extracting Recognition Information from Planning Definition}\label{section:InferringStructures}

In this section, we describe the process through which we extract useful information for plan recognition from a planning domain. 
First, we describe landmark extraction algorithms from the literature, and how we use these algorithms to our approach. 
Second, we show how we classify facts into partitions from planning action descriptions.

\subsection{Extracting Landmarks}

In the planning literature, landmarks~\cite{Hoffmann2004_OrderedLandmarks} are defined as necessary features that must be true at some point in every valid plan to achieve a particular goal. 
Landmarks are often partially ordered according to the sequence in which they must be achieved. 
Hoffman \textit{et al.}~\cite{Hoffmann2004_OrderedLandmarks} define landmarks as follows. 

\begin{definition}[\textbf{Landmark}]\label{def:planLandmark}
\textit{Given a planning instance $\Pi = \langle \Xi, \mathcal{I}, G\rangle$, a formula $L$ is a landmark in $\Pi$ iff $L$ is true at some point along all valid plans that achieve $G$ from $\mathcal{I}$. 
In other words, a landmark is a type of formula (\textit{e.g}, conjunctive formula or disjunctive formula) over a set of facts that must be satisfied at some point along all valid plan executions.}
\end{definition}

For plan recognition problems, landmarks allow us to infer whether a sequence of observations cannot possibly lead to a certain goal. 
In order to extract the landmarks of a planning problem, we use two landmark extraction algorithms from the literature: 1) Hoffman \textit{et al.}~\cite{Hoffmann2004_OrderedLandmarks} to extract conjunctive landmarks; and 2) Porteous and Cresswell~\cite{Porteous_Landmarks_2002} to extract conjunctive and disjunctive landmarks. 
To represent landmarks and their ordering, these algorithms use a tree in which nodes represent landmarks and edges represent necessary prerequisites between landmarks. 
Each node in the tree represents a conjunction of facts that must be true simultaneously at some point during plan execution, and the root node is a landmark representing the goal state. 
These algorithms use a Relaxed Planning Graph (RPG)~\cite{ReachabilityBryceK_2007}, which is a leveled graph that ignores the delete-list effects of all actions, and this way, there are no mutex relations in this graph. 
Once the RPG is built, the algorithm extracts \emph{landmark candidates} by back-chaining from the RPG level in which all facts of the goal state $G$ are possible, and, for each fact $g$ in $G$, checks which facts must be true until the first level of the RPG. 
For example, if fact $B$ is a landmark and all actions that achieve $B$ share $A$ as precondition, then $A$ is a landmark candidate. 
To confirm that a landmark candidate is indeed a landmark, the algorithm builds a new RPG structure by removing actions that achieve this landmark candidate and checks the solvability over this modified problem\footnote{Deciding the solvability of a relaxed planning problem using an RPG structure can be done in polynomial time~\cite{BlumFastPlanning_95}.}, and, if the modified problem is unsolvable, then the landmark candidate is a necessary landmark. 
This means that the actions that achieve the landmark candidate are necessary to solve the original planning problem. 

Hoffman \textit{et al.}~\cite{Hoffmann2004_OrderedLandmarks} proves that the process of generating exactly all landmarks and deciding about their ordering is PSPACE-complete, which is exactly the same complexity of deciding plan existence~\cite{PlanningComplexity_Bylander1994}. 
Nevertheless, most landmark extraction algorithms extract only a subset of landmarks for a given planning instance in order to extract landmarks efficiently. 
In this way, we can monitor landmarks during plan execution to determine which goals a plan is going to achieve and discard candidate goals if some landmarks are not achievable or do not appear as precondition or effect of actions in the observations. 

\subsection{Fact Partitioning}

Pattison and Long~\cite{PattisonGoalRecognition_2010} classify facts into mutually exclusive partitions in order to infer whether certain observations are likely to be goals for goal recognition. 
Their classification relies on the fact that, in some planning domains, predicates may provide additional information that can be extracted by analyzing preconditions and effects in operator definition. 
We use this classification to infer if certain observations are consistent with a particular goal, and if not, we can eliminate a candidate goal. 
We formally define fact partitions in what follows. 

\begin{definition} [\textbf{Strictly Activating}] \label{def:strictlyActivating}
\textit{A fact $f$ is strictly activating if $f \in \mathcal{I}$ and $\forall a \in \mathcal{A}$, such that $f \notin$ \textit{eff}($a$)$^+$ $\cup$ \textit{eff}($a$)$^-$. Furthermore, $\exists a \in \mathcal{A}$, such that $f \in$ \textit{pre}($a$).}
\end{definition}


\begin{definition}[\textbf{Unstable Activating}] \label{def:unstableActivating}
\textit{A fact $f$ is unstable activating if $f \in \mathcal{I}$ and $\forall a \in \mathcal{A}$, $f \notin$ \textit{eff}($a$)$^+$ and $\exists a \in \mathcal{A}, f \in$ \textit{pre}($a$) and $\exists a \in \mathcal{A}$, $f \in$ \textit{eff}($a$)$^-$.}
\end{definition}


\begin{definition}[\textbf{Strictly Terminal}] \label{def:strictlyTerminal}
\textit{A fact $f$ is strictly terminal if $\exists a \in \mathcal{A}$, such that $f \in$ \textit{eff}($a$)$^+$ and $\forall a \in \mathcal{A}$, $f \notin$ \textit{pre}($a$) and $f \notin$ \textit{eff}($a$)$^-$.}
\end{definition}


A \textit{Strictly Activating} fact (Definition~\ref{def:strictlyActivating}) appears as a precondition, and does not appear as add or delete effect in an operator definition. 
This means that unless defined in the initial state, this fact can never be added or deleted by an operator.
An \textit{Unstable Activating} fact (Definition~\ref{def:unstableActivating}) appears as both a precondition and a delete effect in two operator definitions, so once deleted, this fact cannot be re-achieved. 
The deletion of an unstable activating fact may prevent a plan execution to achieve a goal.
A \textit{Strictly Terminal} fact (Definition~\ref{def:strictlyTerminal}) does not appear as a precondition of any operator definition, and once added, cannot be deleted. 
For some planning domains, this kind of fact is the most likely to be in the set of goal facts, because once added in the current state, it cannot be deleted, and remains true until the final state.

The fact partitions that we can extract depend on the planning domain definition. 
For example, from the \textsc{Blocks-World}\footnote{\textsc{Blocks-World} is a classical planning domain where a set of stackable blocks must be re-assembled on a table~\cite{AutomatedPlanning_Book2011}.} domain, it is not possible to extract any fact partitions.
However, it is possible to extract fact partitions from the \textsc{Easy-IPC-Grid}\footnote{\textsc{Easy-IPC-Grid} domain consists of an agent that moves in a grid using keys to open locked locations.} domain, such as \textit{Strictly Activating} and \textit{Unstable Activating} facts. 
Here, we use fact partitions to obtain additional information on fact landmarks. 
For example, consider an \textit{Unstable Activating} fact landmark $L_{ua}$, so that if $L_{ua}$ is deleted from the current state, then it cannot be re-achieved. 
We can trivially determine that goals for which this fact is a landmark are unreachable, because there is no available action that achieves $L_{ua}$ again.

\section{Filtering Candidate Goals from Landmarks in Observations}\label{section:FilteringGoals}

Key to our approach to plan recognition is the ability to filter candidate goals based on the evidence of fact landmarks and partitioned facts in preconditions and effects of observed actions in a plan execution. 
We now present a filtering process that analyzes fact landmarks in preconditions and effects of observed actions, and selects goals, from a set of candidate goals, that have achieved most of their associated landmarks. 

This filtering process is detailed in function \textsc{FilterCandidateGoals} of Algorithm~\ref{alg:filterCandidateGoals}, which takes as input a plan recognition problem $T_{PR}$, which is composed of a planning domain definition $\Xi$, an initial state $\mathcal{I}$, a set of candidate goals $\mathcal{G}$, a set of observed actions $O$, and a filtering threshold $\theta$. 
Our algorithm iterates over the set of candidate goals $\mathcal{G}$, and, for each goal $G$ in $\mathcal{G}$, it extracts and classifies fact landmarks and partitions for $G$ from the initial state $\mathcal{I}$ (Lines~\ref{alg:filter:extractLandmarks}~and~\ref{alg:filter:factPartitioner}). 
We then check whether the observed actions $O$ contain fact landmarks or partitioned facts in either their preconditions or effects. 
At this point, if any \textit{Strictly Activating} facts for the candidate goal $G$ are not in initial state $\mathcal{I}$, then the candidate goal $G$ is no longer achievable and we discard it (Line~\ref{alg:filter:SA}). 
Subsequently, we check for \textit{Unstable Activating} and \textit{Strictly Terminal} facts of goal $G$ in the preconditions and effects of the observed actions $O$, and if we find any, we discard the candidate goal $G$ (Line~\ref{alg:filter:STandUA}).
If we observe no facts from partitions as evidence from the actions in $O$, we move on to checking landmarks of $G$ within the actions in $O$. 
If we observe any landmarks in the preconditions and positive effects of the observed actions (Line~\ref{alg:filter:identifyFactLandmarks}), we compute the percentage of achieved landmarks for goal $G$. 
As we deal with partial observations in a plan execution some executed actions may be missing from the observation, thus whenever we identify a fact landmark, we also infer that its predecessors have been achieved. 
For example, let us consider that the set of fact landmarks to achieve a goal from a state is represented by the following ordered facts: \pred{(at A)} $\prec$ \pred{(at B)} $\prec$ \pred{(at C)} $\prec$ \pred{(at D)}, and we observe just one action during a plan execution, and this observed action contains the~\pred{(at C)} fact landmark as an effect. 
Based on this action from a partial observation, we can infer that the predecessors of \pred{(at C)} have been achieved before the observation, and thus, we also include them as achieved landmarks. 
Given the number of achieved fact landmarks of $G$, we estimate the percentage of fact landmarks that the observed actions $O$ have achieved according to the ratio between the amount of achieved fact landmarks and the total amount of landmarks (Line~\ref{alg:filter:ratio}). 
Finally, after analyzing all candidate goals in $\mathcal{G}$, we return the goals with the highest percentage of achieved landmarks within our filtering threshold $\theta$ (Line~\ref{alg:filter:filterGoals}). 
Note that, if threshold $\theta=0$, the filter returns only the goals with maximum completion, given the observations. 
The threshold gives us flexibility when dealing with incomplete observations and sub-optimal plans, which, when $\theta=0$, may cause some potential goals to be filtered out before we get additional observations. 

\def\gets{:=}

\floatname{algorithm}{Algorithm}
\begin{algorithm}[tb]
    \caption{Filter candidate goals.}
    \textbf{Input:} $\Xi$ $=$ $\langle$$\Sigma$, $\mathcal{A}$$\rangle$ \textit{planning domain}, $\mathcal{I}$ \textit{initial state}, $\mathcal{G}$ \textit{set of candidate goals}, $O$ \textit{observations}, and $\theta$ \emph{threshold}.
    \\\textbf{Output:} \textit{A set of filtered candidate goals $\Lambda_{\mathcal{G}}$ with the highest percentage of achieved landmarks.}
	\label{alg:filterCandidateGoals}
    \begin{algorithmic}[1]
        \Function{FilterCandidateGoals}{$\Xi, \mathcal{I}, \mathcal{G}, O, \theta$}
        \State $\Lambda_{\mathcal{G}}$ := $\langle \rangle$ \Comment{\textit{Map goals to \% of landmarks achieved }.}
        \For{each goal $G$ in $\mathcal{G}$}
			\State $\mathcal{L}_{G}$ := $\Call{ExtractLandmarks}{\Xi, \mathcal{I}, G}$ \label{alg:filter:extractLandmarks}
			\State $\langle F_{sa}, F_{ua}, F_{st} \rangle \gets \Call{PartitionFacts}{\mathcal{L}_{g}}$ \label{alg:filter:factPartitioner} \linebreak\Comment{\textit{$F_{sa}$: Strictly Activating, $F_{ua}$: Unstable Activating, $F_{st}$: Strictly Terminal.}}
			\If{$F_{sa} \cap \mathcal{I} = \emptyset$}\label{alg:filter:SA}
				\State \textbf{continue} \Comment{\textit{Goal $G$ is no longer possible.}}
			\EndIf
			\State $\mathcal{AL}_{G}$ := $\langle$ $\rangle$ \Comment{\textit{Achieved landmarks for $G$.}}
			\For{each observed action $o$ in $O$}
				\If{$(F_{ua} \cup  F_{st}) \subseteq (\textit{pre}(o) \cup \textit{eff}(o)^+ \cup \textit{eff}(o)^-)$} \label{alg:filter:STandUA}
					\State $discardG = $ \textbf{true}
					\State \textbf{break}
				\Else
					\State $L$ := select all fact landmarks $l$ in $\mathcal{L}_{G}$ such that $l$ $\in$ \textit{pre}($o$) $\cup$ \textit{eff}($o$)$^+$ \label{alg:filter:identifyFactLandmarks}
					\State $\mathcal{AL}_{G}$ := $\mathcal{AL}_{G} \cup L$ 
				\EndIf
			\EndFor
			\If{$discardG$} \textbf{break} \Comment{\textit{Avoid computing achieved landmarks for $G$.}}
			\EndIf
			\State $\Lambda_{\mathcal{G}}$ := $\Lambda_{\mathcal{G}} \cup \langle G,\left(\frac{\mid\mathcal{AL}_{G}\mid}{\mid\mathcal{L}_{G}\mid}\right)\rangle$ \Comment{\textit{Percentage of achieved landmarks for $G$.}} \label{alg:filter:ratio}
		\EndFor
		

		\State \textbf{return} {all $G$ s.t $\langle G, v \rangle$ $\in\Lambda_{\mathcal{G}}$ and \newline \hspace*{\algorithmicindent}{\phantom{return }} $v \geq (\max_{v_i}{ \langle G',v_i \rangle \in \Lambda_{\mathcal{G}}})- \theta $}

		\label{alg:filter:filterGoals}
        \EndFunction
    \end{algorithmic}
\end{algorithm}

\begin{figure}[tb]
  \centering
  \includegraphics[width=1\linewidth]{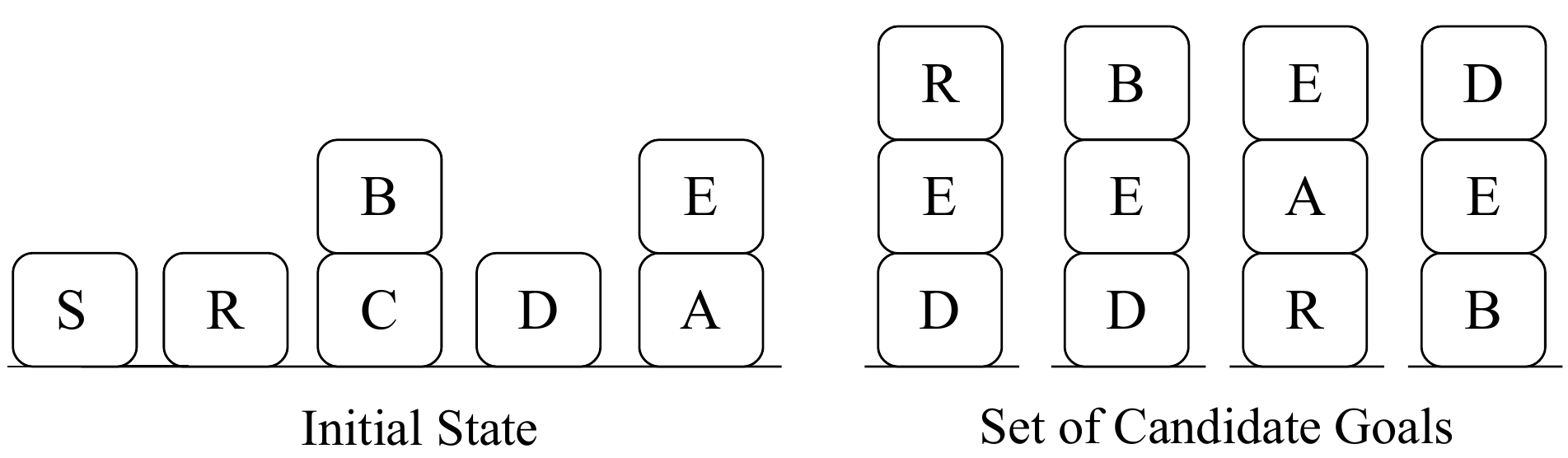}
  \caption{\textsc{Blocks-World} example.}
  \label{fig:blocksExample}
\end{figure}

\begin{figure*}[!ht]
    \begin{minipage}[l]{1.02\columnwidth}
        \centering
        \includegraphics[width=0.97\linewidth]{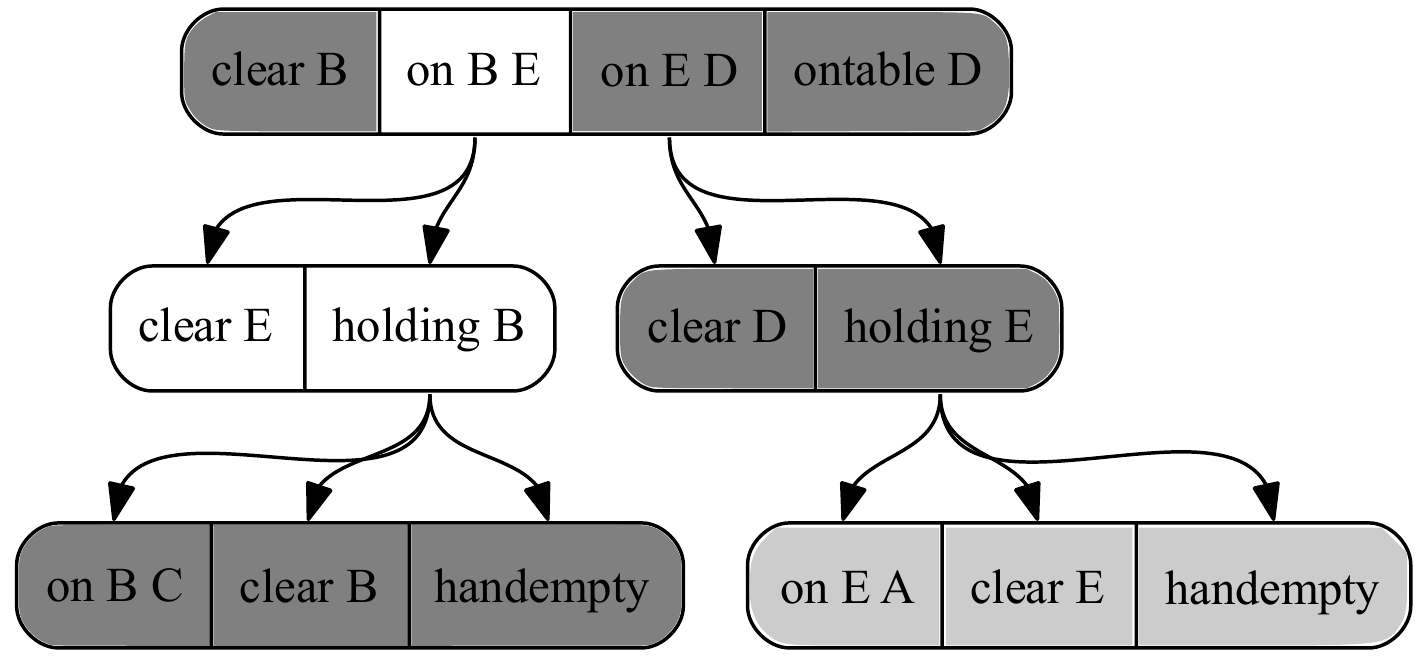}
        \caption{Fact landmarks for the word BED.}\label{fig:BED}
    \end{minipage}
    \begin{minipage}[r]{1.02\columnwidth}
        \centering
        \includegraphics[width=0.97\linewidth]{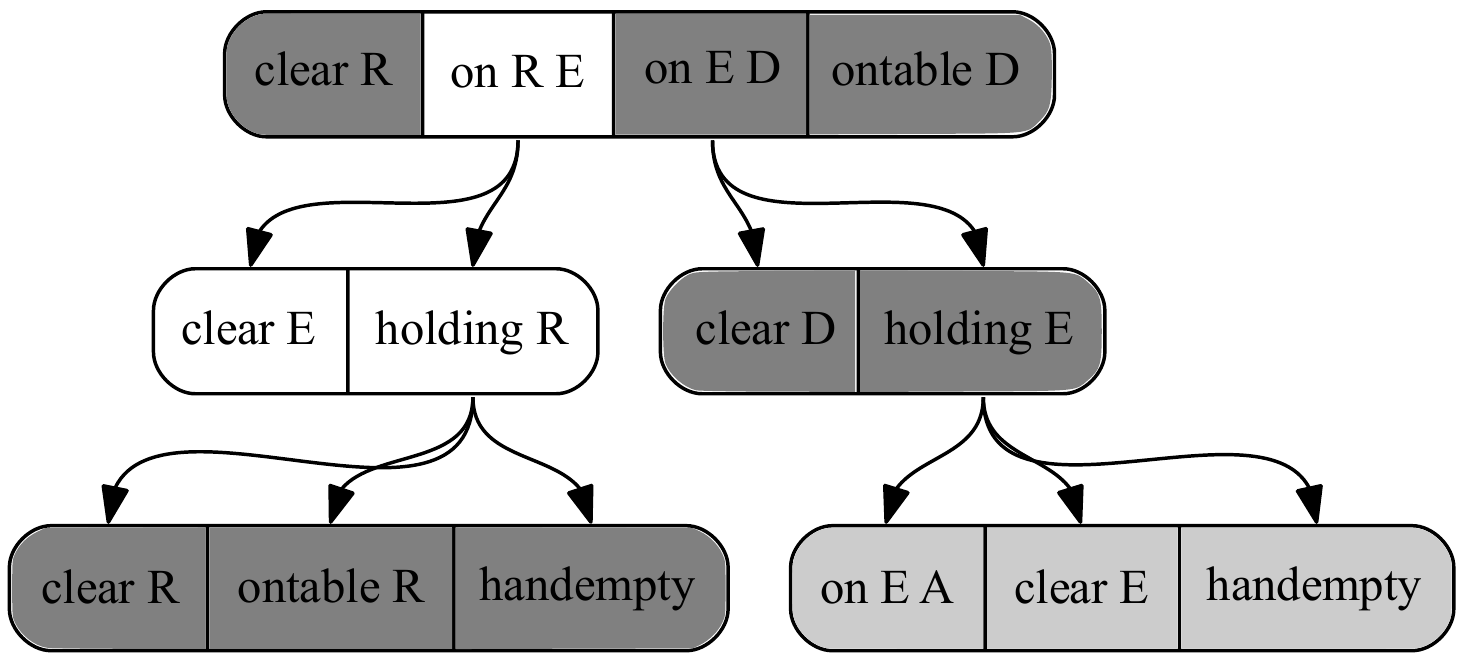}
        \caption{Fact landmarks for the word RED.}\label{fig:RED}
    \end{minipage}
\end{figure*}

As an example of how the algorithm filters a set of candidate goals, consider the \textsc{Blocks-World} example shown in Figure~\ref{fig:blocksExample}, which represents an initial configuration of stackable blocks, as well as set of candidate goals. 
The candidate goals consist of the following stackable words: \pred{BED}, \pred{DEB}, \pred{EAR}, and \pred{RED}. 
Now consider that the following actions have been observed in the plan execution: \pred{(stack E D)} and \pred{(pick-up S)}. 
After filtering the set of candidate goals, we have the following filtered goals for $\theta=0$: \pred{BED} and \pred{RED}. 
Function \textsc{FilterCandidateGoals} returns these goals because the observed action \pred{(stack E D)} has in its preconditions the fact landmarks \pred{(and\,(clear D)\,(holding E))}, and its effects contain \pred{(on E D)}. 
Consequently, from these landmarks, it is possible to infer the evidence for another fact landmark, that is: \pred{(and\,(on E A)\,(clear E)\,(handempty))}. 
This fact landmark is inferred because it must be true before \pred{(clear D)} and \pred{(holding E)}. 
The observed action \pred{(pick-up S)} does not provide any evidence for filtering the set of candidate goals. 
Thus, the estimated percentage of achieved fact landmarks of the filtered candidate goals \pred{BED} and \pred{RED} is 75\%. 
Both of these goals have 8 fact landmarks, and based on the evidence in the observed actions, we infer 6 fact landmarks have been reached, including fact landmarks in the initial state, such as: \pred{(clear B)}, \pred{(ontable D)}, and \pred{(and\,(on B C)\,(clear B)\,(handempty))} for \pred{BED}; and \pred{(clear R)}, \pred{(ontable D)}, and \pred{(and\,(clear R)\,(ontable R)\,(handempty))} for \pred{RED}. 
Regarding goals \pred{EAR} and \pred{DEB}, the observations allow us to conclude that, respectively, 3 and 2 out of 7 and 9 fact landmarks were reached. 
Figures~\ref{fig:BED} and~\ref{fig:RED} show the ordered fact landmarks for the filtered candidate goals \pred{BED} and \pred{RED}. 
Boxes in dark gray show achieved fact landmarks for these goals while boxes in light gray show inferred fact landmarks.

\section{Heuristic Plan Recognition using Landmarks}\label{section:HeuristicPlanRecognition}

We now develop a landmark-based heuristic method that estimates the goal completion of every goal in the set of filtered goals by analyzing the number of achieved landmarks for each goal provided by the filtering process. 
We can now heuristically estimate the goal completion of every goal in the set of filtered goals using the computed landmarks. 
Each candidate goal is composed of sub-goals: atomic facts that are part of a conjunction of facts.
This estimate represents the percentage of sub-goals (atomic facts that are part of a conjunction of facts) in a goal that have been accomplished based on the evidence of achieved fact landmarks in observations. 

Our heuristic method estimates the percentage of completion towards a goal by using the set of achieved fact landmarks provided by the filtering process (Algorithm~\ref{alg:filterCandidateGoals}, Line~\ref{alg:filter:identifyFactLandmarks}). 
We aggregate the percentage of completion of each sub-goal into an overall percentage of completion for all facts in a candidate goal. 
This heuristic, denoted as $h_{prl}$, is computed by the formula below, where $\mathcal{AL}_{g}$ is the number of achieved landmarks from observations of every sub-goal $g$ of the candidate goal $G$, and $\mathcal{L}_{g}$ represents the number of necessary landmarks to achieve every sub-goal $g$ of $G$:

\begin{equation}
\Large
\label{eq:heuristic}
h_{prl}(G) = \left(\frac{\sum_{g \in G} \frac{\mid\mathcal{AL}_{g}\mid}{\mid\mathcal{L}_{g}\mid}}{\mid G \mid}\right)
\end{equation}

Thus, heuristic $h_{prl}(G)$ estimates the completion of a goal $G$ by calculating the ratio between the sum of the percentage of completion for every sub-goal $g \in G$, \textit{i.e}, $\sum_{g \in G} \frac{\mid\mathcal{AL}_{g}\mid}{\mid\mathcal{L}_{g}\mid}$, and the number of sub-goals in $G$.

To exemplify how heuristic $h_{prl}$ estimates goal completion, recall the \textsc{Blocks-World} example from Figure~\ref{fig:blocksExample}. 
For the \pred{BED} goal its sub-goals (shown at the top of Figure~\ref{fig:BED}) are: \pred{(clear B)}, \pred{(on B E)}, \pred{(on E D)}, and \pred{(ontable D)}. 
Based on the observed actions \pred{(stack E D)} and \pred{(pick-up S)}, we conclude that sub-goals \pred{(clear B)} and \pred{(ontable D)} have already been achieved because they are in the initial state, and the observed actions do not delete any of those facts. 
Although fact \pred{(clear B)} in the initial state does not correspond to the final configuration of goal \pred{BED}, we account for this fact in the heuristic calculation, since we consider all observed evidence. 
At this point, our heuristic computes that 50\% of goal \pred{BED} has been accomplished. 
However, for this goal, there is even more information to be considered in order to calculate the percentage of the \pred{BED} goal completion. 
The observed actions have achieved fact landmarks that correspond to the sub-goal \pred{(on E D)}, such as preconditions and effects of the observed action \pred{(stack E D)}, which are: \pred{(and\,(clear D)\,(holding E))}, and \pred{(on E D)}. 
Therefore, we infer that fact landmark \pred{(and\,(on E A)\,(clear E)\,(handempty))} has been achieved, because it must be true before fact landmark \pred{(and\,(clear D)\,(holding E))}. 
For the sub-goal  \pred{(on B E)}, the initial state provides the evidence of the following fact landmark: \pred{(and\,(on B C)\,(clear B)\,(handempty))}.
The observed action \pred{(pick-up S)} does not provide any evidence for the goal \pred{BED}. 
Thus, heuristic $h_{prl}$ estimates that from the evidence of landmarks in the observed actions, the percentage of completion for the goal \pred{BED} is 83.3\%, as follows: \pred{(clear B)} = $\frac{1}{1}$ $+$ \pred{(on B E)} = $\frac{1}{3}$ $+$ \pred{(on E D)} = $\frac{3}{3}$ $+$ \pred{(ontable D)} = $\frac{1}{1}$. 
Note that, by varying the threshold $\theta$ in the filter of Algorithm~\ref{alg:filterCandidateGoals}, we increase the number of candidate goals for which we must compute the heuristic. 
However, since the heuristic is linear on the number of predicates in a goal, increasing the number of candidate goals has virtually no impact in computational complexity.

\section{Landmark-based Plan Recognition}\label{section:landmarkPlanRecognition}

We now bring together the techniques from Sections~\ref{section:FilteringGoals} and \ref{section:HeuristicPlanRecognition} into our landmark-based plan recognition approach that uses the presented filter and heuristic for recognizing goals and plans. 
Our plan recognition approach is detailed in Algorithm~\ref{alg:planRecognition}. 
This algorithm takes as input a plan recognition problem $T_{PR}$, and works in two stages. 
In the first stage, this algorithm filters candidate goals using the filter (Algorithm~\ref{alg:filterCandidateGoals}), which returns the candidate goals with the highest percentage of achieved landmarks within a given threshold $\theta$. 
In the second stage, from the filtered candidates, this algorithm uses our landmark-based heuristic (Equation~\ref{eq:heuristic}) to return the recognized goal(s) by estimating the percentage of completion using the set of achieved fact landmarks provided by the filter.

\floatname{algorithm}{Algorithm}
\begin{algorithm}[ht]
    \caption{Recognize goals and plans using the filtering method and the landmark-based heuristic.} 
    \textbf{Input:} $\Xi$ $=$ $\langle$$\Sigma$, $\mathcal{A}$$\rangle$ \textit{planning domain}, $\mathcal{I}$ \textit{initial state}, $\mathcal{G}$ \textit{set of candidate goals}, $O$ \textit{observations}, and $\theta$ \emph{threshold}.
    \\\textbf{Output:} \textit{Recognized goal(s).}
	\label{alg:planRecognition}
    \begin{algorithmic}[1]
        \Function{Recognize}{$\Xi, \mathcal{I}, \mathcal{G}, O, \theta$}
	        \State $\Lambda_{\mathcal{G}}$ := $\langle \rangle$ \Comment{\textit{Map goals to \% of landmarks achieved}.}
        		\State $\Lambda_{\mathcal{G}}$ := \textsc{FilterCandidateGoals}($\Xi, \mathcal{I}, \mathcal{G}, O, \theta$)
        		\State \textbf{return} $\displaystyle\arg\max_{G \in\Lambda_{\mathcal{G}}} h_{prl}(G)$
        \EndFunction
    \end{algorithmic}
\end{algorithm}

\section{Experiments and Evaluation}\label{section:ExperimentsAndEvaluation}

\begin{table*}[t]
\scriptsize
\centering
\def\arraystretch{0.55}
\begin{tabular}{|ccccc|cc|lc|cc|}
\hline
\multicolumn{1}{|l}{}                                                                            & \multicolumn{1}{l}{}         
& \multicolumn{1}{l}{}                                            
& \multicolumn{1}{l}{}
& \multicolumn{1}{l}{}                                                                     & \multicolumn{2}{|c|}{\textsc{Landmark-based Plan Recognition}}                                                                                                      & \multicolumn{2}{c|}{\textsc{R\&G}}                                                                                                                                            & \multicolumn{2}{c|}{\textsc{Filter + R\&G}}                                                                                                           \\ \hline
\multicolumn{1}{|c|}{{\bf Domain}}                                                               & \multicolumn{1}{c|}{{\bf\hspace*{-1em}$|\mathcal{G}|$\hspace*{-1em}}} & \multicolumn{1}{c|}{{\bf\hspace*{-1em}$|\mathcal{L}|$\hspace*{-1em}}} & {\hspace*{-1em}\bf{\%Obs}\hspace*{-1em}} & {\hspace*{-1em}\bf{$|O|$}\hspace*{-1em}} & \textbf{\begin{tabular}[c]{@{}c@{}}Time\\$\theta$ (0 / 10 / 20 / 30)\end{tabular}}                                                               & \textbf{\begin{tabular}[c]{@{}c@{}}Accuracy\\$\theta$ (0 / 10 / 20 / 30)\end{tabular}}                                                          & \multicolumn{1}{c}{{\bf Time}}                                                                    & {\bf Accuracy}                                                          & {\bf Time}                                                               & {\bf Accuracy}                                                           \\ \hline
\multicolumn{1}{|c|}{{\begin{tabular}[c]{@{}c@{}} {\tiny $\textsc{Blocks-World}$} \\ (855)\end{tabular}}}        & \multicolumn{1}{c|}{20}  & \multicolumn{1}{c|}{15.6}                     & \begin{tabular}[c]{@{}c@{}}10\\ 30\\ 50\\ 70\\ 100\end{tabular} & \begin{tabular}[c]{@{}c@{}}1.1\\ 2.9\\ 4.2\\ 6.5\\ 8.5\end{tabular} & \begin{tabular}[c]{@{}c@{}} 0.99 / 0.100 / 0.105 / 0.111 \\ 0.107 / 0.109 / 0.118 / 0.122 \\ 0.113 / 0.113 / 0.120 / 0.127 \\ 0.138 / 0.139 / 0.141 / 0.148 \\ 0.163 / 0.166 / 0.172 / 0.185 \end{tabular} & \begin{tabular}[c]{@{}c@{}}36.1\% / 38.8\% / 70.0\% / 89.4\% \\ 54.4\% / 61.1\% / 86.1\% / 97.2\% \\ 63.8\% / 83.8\% / 98.3\% / 100.0\% \\ 81.6\% / 94.4\% / 100.0\% / 100.0\% \\ 100.0\% / 100.0\% / 100.0\% / 100.0\% \end{tabular} & \begin{tabular}[c]{@{}l@{}}1.656\\ 1.735\\ 1.836\\ 2.056\\ 2.378\end{tabular}                     & \begin{tabular}[c]{@{}c@{}}83.8\%\\ 90.0\%\\ 97.2\%\\ 98.8\%\\ 100.0\%\end{tabular} & \begin{tabular}[c]{@{}c@{}}0.452\\ 0.458\\ 0.462\\ 0.483\\ 0.494\end{tabular} & \begin{tabular}[c]{@{}c@{}}52.7\%\\ 77.7\%\\ 94.4\%\\ 96.1\%\\ 100.0\%\end{tabular}  \\ \hline

\multicolumn{1}{|c|}{{\begin{tabular}[c]{@{}c@{}} {\tiny $\textsc{Campus}$} \\ (75)\end{tabular}}}           & \multicolumn{1}{c|}{2} & \multicolumn{1}{c|}{8.5}                   & \begin{tabular}[c]{@{}c@{}}10\\ 30\\ 50\\ 70\\ 100\end{tabular} & \begin{tabular}[c]{@{}c@{}}1\\ 2\\ 3\\ 4.4\\ 5.5\end{tabular} & \begin{tabular}[c]{@{}c@{}} 0.038 / 0.039 / 0.042 / 0.044 \\ 0.048 / 0.050 / 0.055 / 0.057 \\ 0.063 / 0.062 / 0.066 / 0.068 \\ 0.060 / 0.060 / 0.063 / 0.065\\ 0.068 / 0.069 / 0.073 / 0.072\end{tabular} & \begin{tabular}[c]{@{}c@{}} 93.3\% / 100.0\% / 100.0\% / 100.0\% \\ 100.0\% / 100.0\% / 100.0\% / 100.0\% \\ 93.3\% / 100.0\% / 100.0\% / 100.0\% \\ 100.0\% / 100.0\% / 100.0\% / 100.0\% \\ 100.0\% / 100.0\% / 100.0\% / 100.0\% \end{tabular} & \begin{tabular}[c]{@{}l@{}}0.083\\ 0.091\\ 0.105\\ 0.112\\ 0.126\end{tabular}                     & \begin{tabular}[c]{@{}c@{}}100.0\%\\ 100.0\%\\ 100.0\%\\ 100.0\%\\ 100.0\%\end{tabular}    & \begin{tabular}[c]{@{}c@{}}0.090\\ 0.089\\ 0.092\\ 0.095\\ 0.097\end{tabular} & \begin{tabular}[c]{@{}c@{}}100.0\%\\ 100.0\%\\ 100.0\%\\ 100.0\%\\ 100.0\%\end{tabular}  \\ \hline

\multicolumn{1}{|c|}{{\begin{tabular}[c]{@{}c@{}} {\tiny $\textsc{Easy-IPC-Grid}$} \\ (465)\end{tabular}}}       & \multicolumn{1}{l|}{7.5} & \multicolumn{1}{c|}{11.3}                                     & \begin{tabular}[c]{@{}c@{}}10\\ 30\\ 50\\ 70\\ 100\end{tabular} & \begin{tabular}[c]{@{}c@{}}1.8\\ 4.3\\ 6.9\\ 9.8\\ 13.3\end{tabular} & \begin{tabular}[c]{@{}c@{}}  0.585 / 0.588 / 0.609 / 0.623 \\ 0.597 / 0.600 / 0.614 / 0.644 \\ 0.608 / 0.609 / 0.627 / 0.656 \\ 0.629 / 0.628 / 0.661 / 0.715 \\ 0.630 / 0.632 / 0.685 / 0.759 \end{tabular} & \begin{tabular}[c]{@{}c@{}} 82.2\% / 85.5\% / 97.7\% / 100.0\% \\ 86.6\% / 93.3\% / 97.7\% / 100.0\% \\ 94.4\% / 97.7\% / 97.7\% / 100.0\% \\ 95.5\% / 98.8\% / 98.8\% / 100.0\%  \\ 100.0\% / 100.0\% / 100.0\% / 100.0\% \end{tabular} & \multicolumn{1}{c}{\begin{tabular}[c]{@{}c@{}}1.206\\ 1.291\\ 1.306\\ 1.715\\ 2.263\end{tabular}} & \begin{tabular}[c]{@{}c@{}}97.7\%\\ 98.8\%\\ 98.8\%\\ 100.0\%\\ 100.0\%\end{tabular}  & \begin{tabular}[c]{@{}c@{}}0.770\\ 0.790\\ 0.860\\ 0.932\\ 1.091\end{tabular} & \begin{tabular}[c]{@{}c@{}}97.7\%\\ 98.8\%\\ 100.0\%\\ 100.0\%\\ 100.0\%\end{tabular} \\ \hline

\multicolumn{1}{|c|}{{\begin{tabular}[c]{@{}c@{}} {\tiny $\textsc{Intrusion-Detection}$} \\ (465)\end{tabular}}} & \multicolumn{1}{c|}{15}  & \multicolumn{1}{c|}{16}                                        & \begin{tabular}[c]{@{}c@{}}10\\ 30\\ 50\\ 70\\ 100\end{tabular} & \begin{tabular}[c]{@{}c@{}}1.9\\ 4.5\\ 6.7\\ 9.5\\ 13.1\end{tabular}                  & \begin{tabular}[c]{@{}c@{}}0.197 / 0.200 / 0.211 / 0.233 \\ 0.214 / 0.219 / 0.227 / 0.241 \\ 0.218 / 0.221 / 0.246 / 0.269 \\ 0.219 / 0.223 / 0.258 / 0.274\\ 0.277 / 0.281 / 0.303 / 0.325 \end{tabular} & \begin{tabular}[c]{@{}c@{}} 76.4\% / 96.6\% / 100.0\% / 100.0\% \\ 94.4\% / 100.0\% / 100.0\% / 100.0\% \\ 100.0\% / 100.0\% / 100.0\% / 100.0\%\\ 100.0\% / 100.0\% / 100.0\% / 100.0\%\\ 100.0\% / 100.0\% / 100.0\% / 100.0\% \end{tabular}  & \begin{tabular}[c]{@{}l@{}}1.130\\ 1.142\\ 1.203\\ 1.482\\ 1.567\end{tabular}                     & \begin{tabular}[c]{@{}c@{}}98.8\%\\ 100.0\%\\ 100.0\%\\ 100.0\%\\ 100.0\%\end{tabular}    & \begin{tabular}[c]{@{}c@{}}0.506\\ 0.521\\ 0.531\\ 0.568\\ 0.566\end{tabular} & \begin{tabular}[c]{@{}c@{}}98.8\%\\ 100.0\%\\ 100.0\%\\ 100.0\%\\ 100.0\%\end{tabular}   \\ \hline

\multicolumn{1}{|c|}{{\begin{tabular}[c]{@{}c@{}} {\tiny $\textsc{Kitchen}^\dagger$} \\ (75)\end{tabular}}}           & \multicolumn{1}{c|}{3} & \multicolumn{1}{c|}{5}      & \begin{tabular}[c]{@{}c@{}}10\\ 30\\ 50\\ 70\\ 100\end{tabular} & \begin{tabular}[c]{@{}c@{}}1.3\\ 3.5\\ 4\\ 5\\ 7.4\end{tabular} & \begin{tabular}[c]{@{}c@{}}0.003 / 0.003 / 0.002 / 0.004 \\ 0.003 / 0.004 / 0.005 / 0.005 \\ 0.004 / 0.004 / 0.006 / 0.006 \\ 0.006 / 0.007 / 0.007 / 0.008\\ 0.007 / 0.008 / 0.008 / 0.009\end{tabular} & \begin{tabular}[c]{@{}c@{}} 93.3\% / 100.0\% / 100.0\% / 100.0\%\\ 93.3\% / 100.0\% / 100.0\% / 100.0\% \\ 93.3\% / 100.0\% / 100.0\% / 100.0\%\\ 93.3\% /  93.3\% / 100.0\% / 100.0\% \\ 100.0\% / 100.0\% / 100.0\% / 100.0\% \end{tabular} & \begin{tabular}[c]{@{}c@{}}0.099\\ 0.111\\ 0.112\\ 0.111\\ 0.118\end{tabular}                     & \begin{tabular}[c]{@{}c@{}}100.0\%\\ 100.0\%\\ 100.0\%\\ 100.0\%\\ 100.0\%\end{tabular}    & \begin{tabular}[c]{@{}c@{}}0.093\\ 0.107\\ 0.111\\ 0.110\\ 0.112\end{tabular} & \begin{tabular}[c]{@{}c@{}}100.0\%\\ 100.0\%\\ 100.0\%\\ 100.0\%\\ 100.0\%\end{tabular}  \\ \hline

\multicolumn{1}{|c|}{{\begin{tabular}[c]{@{}c@{}} {\tiny $\textsc{Logistics}$} \\ (465)\end{tabular}}}           & \multicolumn{1}{c|}{10} & \multicolumn{1}{c|}{18.7}     & \begin{tabular}[c]{@{}c@{}}10\\ 30\\ 50\\ 70\\ 100\end{tabular} & \begin{tabular}[c]{@{}c@{}}2\\ 5.9\\ 9.5\\ 13.4\\ 18.7\end{tabular}                    & \begin{tabular}[c]{@{}c@{}}0.441 / 0.449 / 0.455 / 0.458 \\ 0.447 / 0.452 / 0.461 / 0.466 \\ 0.457 / 0.469 / 0.474 / 0.488 \\ 0.474 / 0.481 / 0.490 / 0.497\\ 0.498 / 0.505 / 0.513 / 0.522\end{tabular} & \begin{tabular}[c]{@{}c@{}} 73.3\% / 96.6\% / 100.0\% / 100.0\% \\ 88.7\% / 100.0\% / 100.0\% / 100.0\% \\ 96.6\% / 100.0\% / 100.0\% / 100.0\%\\ 100.0\% / 100.0\% / 100.0\% / 100.0\% \\ 100.0\% / 100.0\% / 100.0\% / 100.0\% \end{tabular} & \begin{tabular}[c]{@{}l@{}}1.125\\ 1.195\\ 1.248\\ 1.507\\ 1.984\end{tabular}                     & \begin{tabular}[c]{@{}c@{}}100.0\%\\ 100.0\%\\ 98.8\%\\ 100.0\%\\ 100.0\%\end{tabular}    & \begin{tabular}[c]{@{}c@{}}0.615\\ 0.663\\ 0.712\\ 0.786\\ 0.918\end{tabular} & \begin{tabular}[c]{@{}c@{}}98.8\%\\ 100.0\%\\ 98.8\%\\ 100.0\%\\ 100.0\%\end{tabular}  \\ \hline
\end{tabular}
\caption{Comparison and experimental results of our landmark-based approach against Ramirez and Geffner~\protect\cite{RamirezG_IJCAI2009} approach. R\&G denotes their plan recognition approach and Filter + R\&G denotes the same approach but using our filtering algorithm. For the experiments with the \textsc{Kitchen} domain we use disjunctive landmarks$^\dagger$.}
\label{tab:ExperimentalResultsPlanRecognition}
\end{table*}

In this section, we describe the experiments and evaluation we carried out on our landmark-based plan recognition approach against state-of-the-art techniques.
For experiments, we use six domains from datasets provided by Ram{\'{\i}}rez and Geffner~\cite{RamirezG_IJCAI2009,RamirezG_AAAI2010}, comprising hundreds of problems\footnote{\url{https://goo.gl/gLF6wB}}. 
We summarize these domains as follows. 

\begin{itemize}
	\item \textsc{Blocks-world} domain consists of a set of blocks, a table, and a robot hand. Blocks can be stacked on top of other blocks or on the table. A block that has nothing on it is clear. The robot hand can hold one block or be empty. The goal is to find a sequence of actions that achieves a final configuration of blocks;
	\item \textsc{Campus} domain consists of finding what activity is being performed by a student from his observations on campus environment;
	\item \textsc{Easy-IPC-Grid} domain consists of an agent that moves in a grid from connected cells to others by transporting keys in order to open locked locations;
	\item \textsc{Intrusion-Detection} represents a domain where a hacker tries to access, vandalize, steal information, or  perform a combination of these attacks on a set of servers;
	\item \textsc{Kitchen} is a domain that consists of home-activities, in which the goals can be preparing dinner, breakfast, among others; and
	\item \textsc{Logistics} is a domain which models cities, and each city contains locations. These locations are airports. For transporting packages between locations, there are trucks and airplanes. Trucks can drive between cities. Airplanes can fly between airports. The goal is to get and transport packages from locations to other locations.
\end{itemize}

These domains contain hundreds of plan recognition problems, \textit{i.e}, a domain description as well as an initial state, a set of candidate goals $\mathcal{G}$, a hidden goal $G$ in $\mathcal{G}$, and an observation sequence $O$.
An observation sequence contains actions that represent an optimal plan or sub-optimal plan that achieves a hidden goal $G$, and this observation sequence can be full or partial. 
A full observation sequence represents the whole plan for a hidden goal $G$, \textit{i.e}, 100\% of the actions having been observed. 
A partial observation sequence represents a plan for a hidden goal $G$ with 10\%, 30\%, 50\%, or 70\% of its actions having been observed. Our experiments use two metrics, the accuracy of the recognition and the speed to recognize a goal. 
We compare our approach to two other approaches: the approach of Ram{\'{\i}}rez and Geffner~\cite{RamirezG_IJCAI2009}, more specifically, we use their faster and most accurate approach; as well as a combination of their approach and our filter.

\begin{figure*}[t]
\begin{minipage}[t]{0.5\linewidth}
    \includegraphics[width=0.715\linewidth, angle=270]{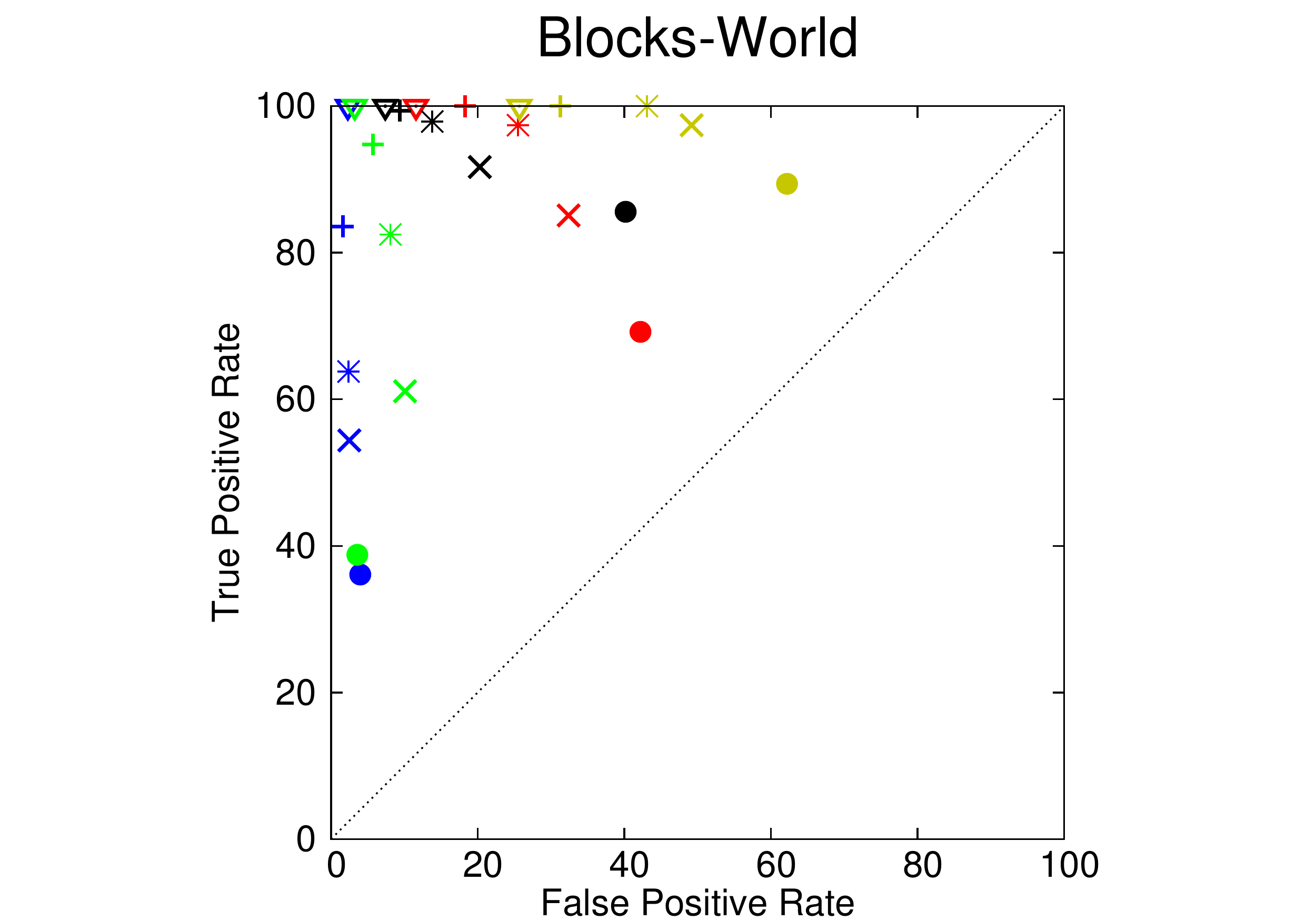}
	\caption{ROC curve for the \textsc{Blocks-World} domain.}
	\label{fig:rocBlocks}
\end{minipage}
\begin{minipage}[t]{0.5\linewidth}
    \includegraphics[width=0.715\linewidth, angle=270]{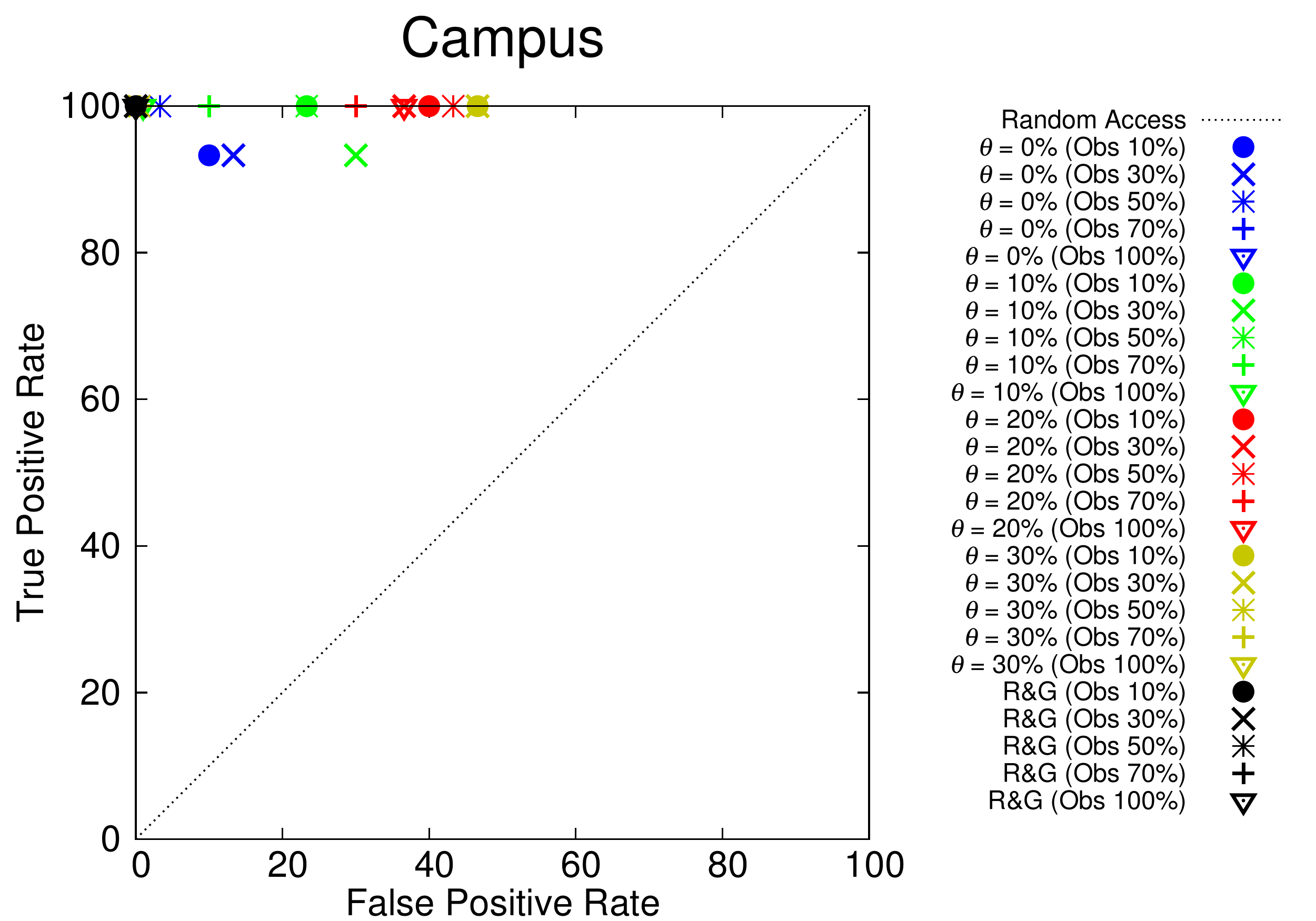}
	\caption{ROC curve for the \textsc{Campus} domain.}
	\label{fig:rocCampus}
\end{minipage}
\end{figure*}

\begin{figure*}[t]
\begin{minipage}[t]{0.5\linewidth}
    \includegraphics[width=0.715\linewidth, angle=270]{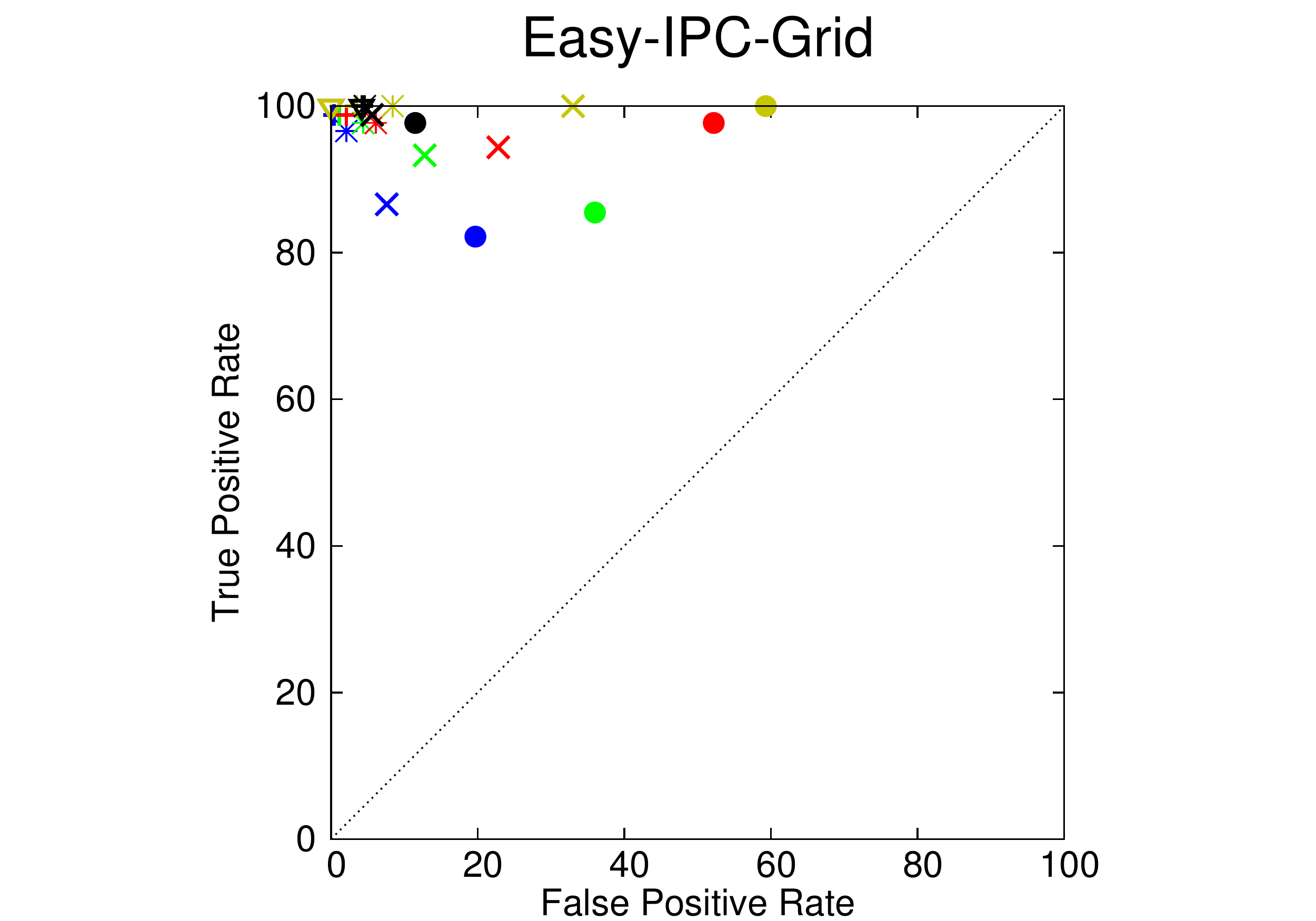}
	\caption{ROC curve for the \textsc{Easy-IPC-Grid} domain.}
	\label{fig:rocGrid}
\end{minipage}
\begin{minipage}[t]{0.5\linewidth}
    \includegraphics[width=0.715\linewidth, angle=270]{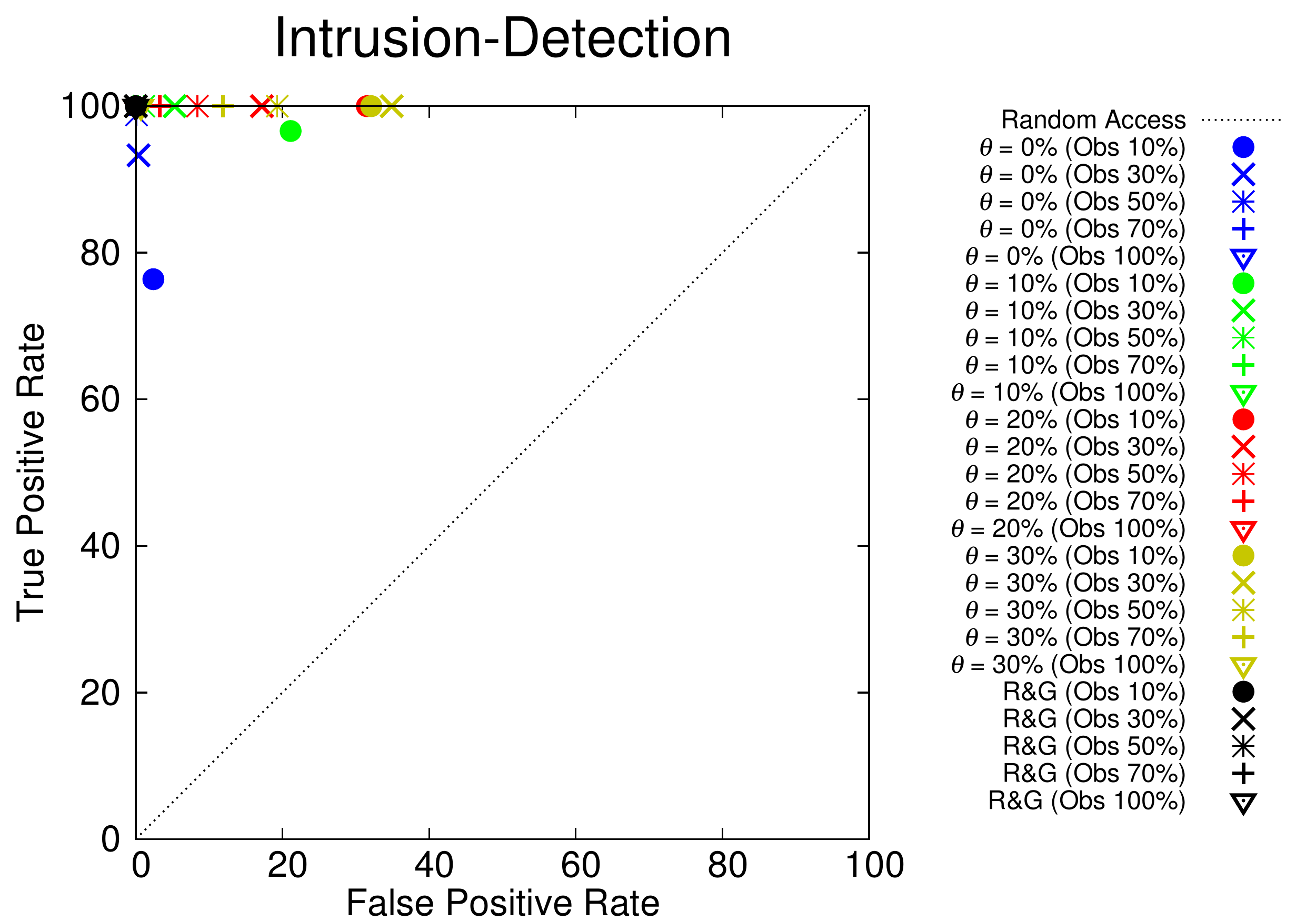}
	\caption{ROC curve for the \textsc{Intrusion-Detection} domain.}
	\label{fig:rocID}
\end{minipage}
\end{figure*}

\begin{figure*}[t]
\begin{minipage}[t]{0.5\linewidth}
    \includegraphics[width=0.715\linewidth, angle=270]{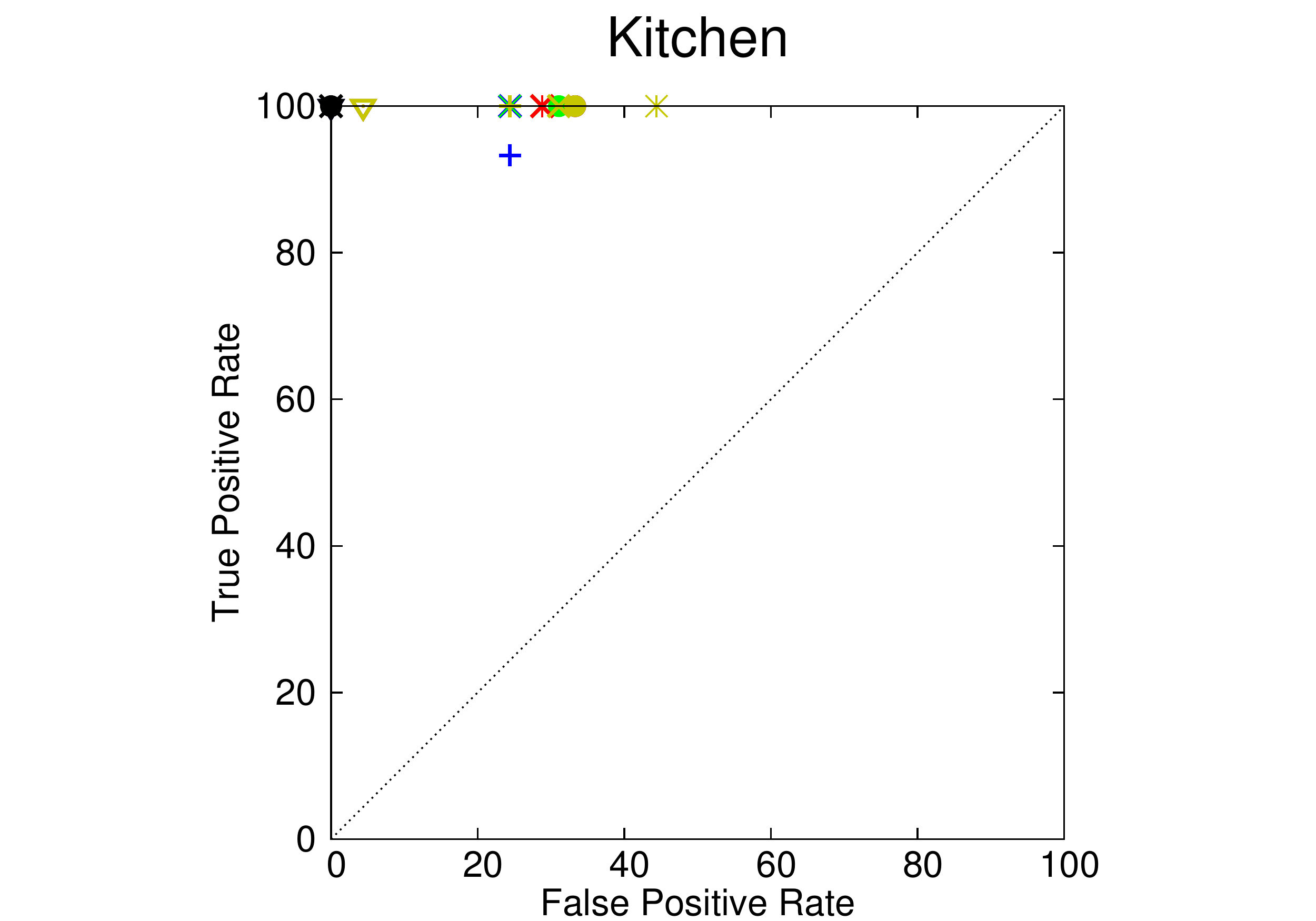}
	\caption{ROC curve for the \textsc{Kitchen} domain.}
	\label{fig:rocKitchen}
\end{minipage}
\begin{minipage}[t]{0.5\linewidth}
    \includegraphics[width=0.715\linewidth, angle=270]{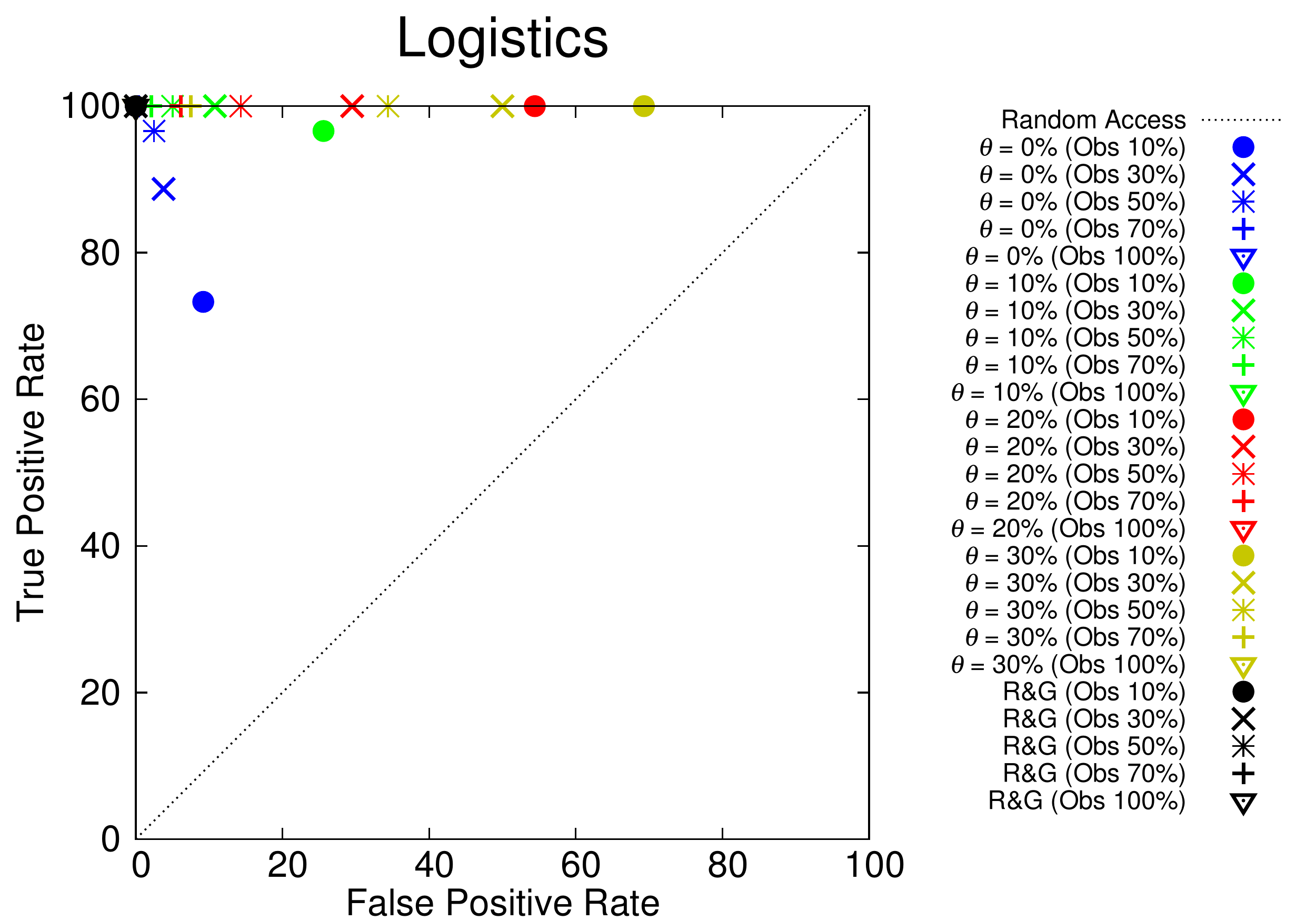}
	\caption{ROC curve for the \textsc{Logistics} domain.}
	\label{fig:rocLogistics}
\end{minipage}
\end{figure*}

For evaluation, we use the accuracy metric (true positive rate), which represents how well a hidden goal is recognized from a set of possible goals for a given plan recognition problem; as well as recognition time (in seconds), which represents how long it takes for a hidden goal to be recognized given a plan recognition problem. In the $\textsc{Blocks-World}$ domain, the accuracy metric measures how well these approaches recognize, from observations, the word that is being assembled. Regarding $\textsc{Campus}$ domain, we aim to accurate how well these approaches recognize the activity is being performed by an observed student. For the $\textsc{Easy-IPC-Grid}$ domain, how accurate these approaches recognize the cell where keys are being to transported by the observed agent. With regard to $\textsc{Intrusion-Detection}$ domain, how accurate these approaches recognize the type of attack and servers that are being hacked observations.
In the \textsc{Kitchen} domain, we aim to accurate how well these approaches recognize the meal is being prepared. For $\textsc{Logistics}$ domain, how accurate these approaches recognize the location where the packages are being transported from observations.
Besides the accuracy metric, we use the Receiver Operating Characteristic (ROC), which is called ROC curve. ROC curve shows graphically the performance of a binary classifier system by evaluating true positive rate against the false positive rate at various threshold settings (in this paper we evaluate plan recognition approaches). More specifically, we use the ROC curve to compare not only true positive predictions (\textit{i.e}, accuracy), but also to compare the false positive ratio of the experimented plan recognition approaches.

Table~\ref{tab:ExperimentalResultsPlanRecognition} compares the results for the three plan recognition approaches, showing the total number of plan recognition problems used under each domain name. 
For each domain we show the number of candidate goals $|\mathcal{G}|$ and varying percentages of the plan that is actually observed, as well as the average number of observed actions per problem $|O|$.
Note that, for partial observations, random observed actions are removed (up to the set percentage), but the order is maintained. 
$|\mathcal{L}|$ denotes the average number of fact landmarks extracted for each domain.
For each approach, we compute the time to recognize the hidden goal (seconds), given the observations, and the accuracy with which the approaches correctly infer the goal. 
For our landmark-based plan recognition approach, we show the accuracy under different filtering thresholds (0\%, 10\%, 20\% and 30\%). If threshold $\theta=0$, our approach does not give any flexibility for filtering candidate goals, returning only the goals with the highest percentage of achieved landmarks. Each row of this table shows the observability (\% Obs) and averages of the number of candidate goals $|\mathcal{G}|$, the number of observed actions $|O|$, recognition time, and accuracy. 
From this table, it is possible to see that our landmark-based plan recognition approach is both faster and more accurate than Ram{\'{\i}}rez and Geffner~\cite{RamirezG_IJCAI2009}, and, when we combine their algorithm with our filter, the resulting approach gets a substantial speedup. Importantly, as we increase the threshold, our plan recognition approach quickly surpasses the state of the art in all domains tested. 
We note that when measuring time to recognition using our filter we also include the time to compute landmarks, so that landmark computation is performed online (\textit{i.e}, during the process of plan recognition). 
Thus, even if this computation has a complex upper bound, in our experience, computing landmarks (especially conjunctive ones) is very fast.

Table~\ref{tab:ExperimentalResultsPlanRecognition} shows that both our landmark-based plan recognition approach and Ram{\'{\i}}rez and Geffner's~\cite{RamirezG_IJCAI2009} yield near perfect accuracy for recognizing goals and plans for all planning domains. 
However, by using the ROC curve we highlight the trade-off between true positive results and false positive results for these plan recognition approaches.
Figures~\ref{fig:rocBlocks}-\ref{fig:rocLogistics} show the ROC curve for the six planning domains we use. 
In the ROC curve, the diagonal line in (Random Access) represents a random guess to recognize a goal from observations. 
This diagonal line divides the ROC space, in which points above the diagonal represent good classification results (better than random), whereas points below the line represent poor results (worse than random). 
The best possible (perfect) prediction for recognizing goals must be a point in the upper left corner (\textit{i.e}, coordinate \textit{x} = 0 and \textit{y} = 100) in the ROC space. 
Thus, the closer a plan recognition approach (point) gets to the upper left corner, the better it is for recognizing goals and plans. 
Blue, green, red, and yellow points with five different symbols represent our plan recognition approach varying the use of the threshold (0\%, 10\%, 20\% and 30\%). 
These five different symbols represent the percentage of observability (10\%, 30\%, 50\%, 70\% and 100\%) with regard to the observed plan. 
Black points represent Ram{\'{\i}}rez and Geffner's~\cite{RamirezG_IJCAI2009} approach (R\&G). 
According to the ROC curve in Figures~\ref{fig:rocCampus},~\ref{fig:rocID},~\ref{fig:rocKitchen}, and~\ref{fig:rocLogistics} we see that all variation (using different thresholds) of our plan landmark-based recognition approach yield good (sometimes perfect) predictions for recognizing goals and plans, in contrast to R\&G, which is near-perfect in these four domains. 
Figure~\ref{fig:rocBlocks} shows that the results for the \textsc{Blocks-World} are quite scattered in the ROC curve, so recognizing goals and plans in this domain is difficult. 
Nevertheless, it possible to see that our plan recognition is not only competitive (using the thresholds between 10\% and 20\%) with R\&G with superior accuracy, but also at least 8.75 orders of magnitude faster than R\&G. 
Finally, Figure~\ref{fig:rocGrid} shows that in the \textsc{Easy-IPC-Grid} domain our approach is very competitive with R\&G, again with consistently higher accuracy, but also is near perfect for false positives, surpassing R\&G by using different thresholds. 

\section{Related Work}\label{section:RelatedWork}

Ram\'{i}rez and Geffner~\cite{RamirezG_IJCAI2009} propose planning approaches for plan recognition, and instead of using plan-libraries, they model the problem as a planning domain theory with respect to a known set of goals. 
Their work uses a heuristic, an optimal and modified sub-optimal planner to determine the distance to every goal in a set of goals after an observation. 
Follow-up work~\cite{RamirezG_AAAI2010} proposes a probabilistic plan recognition approach using off-the-shelf planners. 
These approaches yield high accuracy in most domains, however, this accuracy is lower than in our threshold-based approaches, and their time to recognition ranges from twice slower to up to an order of magnitude slower. 
Pattison and Long~\cite{PattisonGoalRecognition_2010} propose IGRAPH (AUTOmatic Goal Recognition with A Planning Heuristic), a probabilistic heuristic-based goal recognition over planning domains. 
IGRAPH uses heuristic estimation and domain analysis to determine which goals an agent is pursuing.
Although we adapt their fact partitions, their problem definition is formally different than ours, preventing direct comparison. In~\cite{GoalRecognitionDesign_Keren2014}, Keren~\emph{et al.} present an alternate view regarding the goal and plan recognition problem. This work uses planning techniques to assist in the design of goal and plan recognition problems.
Most recently, E.-Martín \emph{et al.}~\cite{NASA_GoalRecognition_IJCAI2015} propose a planning-based plan recognition approach that propagates cost and interaction information in a plan graph, and uses this information to estimate goal probabilities over the set of candidate goals.
Although our landmark-based plan recognition approach has no probabilistic interpretation, the accuracy of our approach seems to be higher in the same domains. 

\section{Conclusion}\label{section:Conclusion}

We have developed an approach for plan recognition that relies on planning landmarks and a new heuristic based on these landmarks. 
Landmarks provide key information about what cannot be avoided to achieve a goal, and we show that landmarks can be used efficiently for very accurate plan recognition. 
We have shown empirically that our approach yields not only superior accuracy results but also substantially faster recognition times for all domains used in evaluating against the state of the art~\cite{RamirezG_IJCAI2009} at varying observation completeness levels. 

Our experiments show that in at least one domain, disjunctive landmarks have a positive effect on accuracy with minimal effect on recognition time, whereas in some domains, these landmarks are either not present or yield almost no gain in accuracy at substantial loss of speed. 
Knowledge of the domains leads us to believe that disjunctive landmarks are most useful in domains in which we assume that observed plans are just sub-optimal, such as the \textsc{Kitchen} domain.
Conversely, disjunctive landmarks slow down recognition in domains in which there are multiple mutually exclusive plans towards the same goal, such as the \textsc{Easy-IPC-Grid} domain in which the agent moves in a grid. 

We intend to explore multiple avenues for future work. 
First, we aim to evaluate other planning techniques, such as heuristics and symmetries in classical planning~\cite{Heuristics2015}. Second, we intend to explore other landmark extraction algorithms to obtain additional information from planning domains, such as temporal landmarks~\cite{Landmarks2015}. 
Third, we aim to model a probability interpretation to the observed landmarks and compare the probability results (and probabilistic accuracy) to the recent work of E.-Martín \emph{et al.}~\cite{NASA_GoalRecognition_IJCAI2015}. 
Finally, for domains with goals with intersecting landmarks, we can use measures of information gain to weigh observations to help break ties when multiple goals are left after the filter.
Given the computational complexity of landmark extraction in the general case, we aim to theoretically analyze the tradeoff between landmark completeness and runtime efficiency. 


\bibliography{planrecognition-ecai}
\end{document}